\documentclass[12pt]{article}
\usepackage[utf8]{inputenc}

\usepackage{amsmath}
\usepackage{amssymb}
\usepackage{amsthm,lineno}
\usepackage{graphicx}
\usepackage{esint}
\usepackage{float}
\usepackage{indentfirst}
\usepackage{amsfonts}
\usepackage{mathrsfs}
\usepackage{graphicx}
\usepackage{booktabs}
\usepackage{subcaption}
\usepackage{bm}
\usepackage{bbm}
\usepackage{wrapfig, enumitem}
\usepackage[colorlinks = true,citecolor=blue]{hyperref}
\usepackage{color}
\usepackage{mathtools}
\usepackage[square,numbers]{natbib}
\usepackage{algorithm}
\usepackage{algorithmic}
\usepackage[a4paper,margin=1in]{geometry}
\usepackage[toc,page]{appendix}
\usepackage{comment}
\usepackage{diagbox}
\usepackage{makecell}
\usepackage{threeparttable}
\usepackage{multirow}
\usepackage{titlesec}
\usepackage{tikz}

\titlespacing*{\section}{0pt}{3pt plus 3pt minus 1pt}{3pt plus 3pt minus 1pt}
\titlespacing*{\subsection}{0pt}{3pt plus 3pt  minus 1pt}{3pt plus 3pt minus 1pt}
\newtheorem{theorem}{Theorem}

\newtheorem{proposition}[theorem]{Proposition}

\newtheorem{definition}{Definition}

\newcommand{\E}{\mathbb{E}}
\newcommand{\bs}{\boldsymbol}

\newcommand{\R}{\mathcal{R}}

\newcommand{\ds}{\displaystyle}

\newcommand{\w}{\bs w}

\title{Efficient Reinforcement Learning  from Demonstration Using Local Ensemble and Reparameterization with Split and Merge of Expert Policies}
\author{Wang Yu\footnote{ywang50@nd.edu}, Fang Liu\footnote{fliu2@nd.edu}}
\date{\textit{Applied Computational Mathematics and Statistics} \\
\textit{University of Notre Dame}\\
Notre Dame, IN 46503, USA \\
}

\begin{document}
\maketitle
\begin{abstract}
The current work on reinforcement learning (RL) from demonstrations 
often assumes the demonstrations are samples from an optimal policy, an unrealistic assumption in practice. When demonstrations are generated by  sub-optimal policies or have sparse state-action pairs, policy learned from  sub-optimal demonstrations may mislead an agent with incorrect or non-local action decisions. We propose a new method called Local Ensemble and Reparameterization with Split and Merge of expert policies (LEARN-SAM) to improve efficiency and make better use of the  sub-optimal demonstrations. First, LEARN-SAM employs a new concept, the $\lambda$-function, based on a discrepancy measure between the current state to demonstrated states to ``localize'' the weights of the expert policies during learning. Second, LEARN-SAM employs a split-and-merge (SAM) mechanism by separating the helpful parts in each expert demonstration and regrouping them into new expert policies to use the demonstrations selectively. Both the $\lambda$-function and SAM mechanism help boost the learning speed. Theoretically, we prove the invariant property of reparameterized policy before and after the SAM mechanism, providing theoretical guarantees for the convergence of the employed policy gradient method. We demonstrate the superiority of the LEARN-SAM method and its robustness with varying demonstration quality and sparsity in six experiments on complex continuous control problems of low to high dimensions, compared to existing methods on RL from demonstration.

\textbf{keyword}: reinforcement Learning from demonstration (RLfD), ensemble,  sub-optimal demonstrations, invariant, localization, reparameterization, optimization, policy gradient

\end{abstract}

%%%%%%%%%%%
\newpage
\section{Introduction}\vspace{-3pt}
\subsection{Background}\vspace{-3pt}
Reinforcement Learning (RL) is a machine learning paradigm where an agent navigates toward the optimal decision-making policy, maximizing the cumulative rewards along the way. RL is shown to be effective and well-suited for a variety of tasks such as robotic control, video games, self-driving, and online recommendation systems, among others. \cite{ mahmood2008adaptive,peters2003reinforcement,sallab2017deep,silver2018general,sutton2018reinforcement} In complex tasks with sparse significant rewards, the agent often takes an excruciatingly long time to collect these rewards, especially if a RL procedure is designed with a \emph{tabula rasa} approach \cite{brys2015policy}. Methods that incorporate prior knowledge have been proposed to speed up learning, among which, RL from demonstration (RLfD) is a mainstream approach, such as reward shaping, imitation learning, and inverse RL \cite{schaal1996learning, russell1998learning,  ng2000algorithms, argall2009survey,  brys2015reinforcement,   ho2016generative, wang2017improving, ramirez2021model}.  

Early techniques of RLfD typically extract prior knowledge by pre-training expert policies with supervised learning methods \cite{silver2016mastering, cruz2017pre, ajay2020opal, nair2020awac, singh2020parrot} or pushing them into memory replay buffer for value evaluation \cite{hester2018deep}. Once a expert policy is trained, it will be used in the subsequent RL as is, without differentiating how helpful it is in assisting the agent to learn or adapting how it can assist  in different states. Recent RLfD methods solve these problems by using reward shaping, applying imitation learning, or leveraging the hierarchical RL framework \cite{brys2015reinforcement, brys2015policy, kang2018policy,  nair2018overcoming, pomerleau1988alvinn, hussein2017imitation, jing2020reinforcement, sutton1999between, cao2012bayesian, daniel2016probabilistic, barto2003recent}. 

Many existing techniques assume the demonstrations are sampled from an optimal policy, and some can deal with demonstrations that are   sub-optimal in the sense that they may be noisy, low-quality or have data sparsity issue\cite{brys2015reinforcement,jing2020reinforcement}, which is the more realistic case. The quality of demonstrations, often measured by the aggregated reward when they are sampled, may degrade with noise from the environment or when the expert policies are immature. The sparsity of demonstrations refers to low numbers of trajectories included in demonstrations or insufficient samples of state-actions pairs in a demonstration. Though the existing methods on RLfD may still be able to work with sub-optimal demonstrations, they can get stuck easily in local optima or be misled by wrong decisions from low-quality demonstrations, leading to performance potentially worse than vanilla learning.  %A policy that imitates the behavior of demonstrated decisions can be obtained by alternating between solving a RL problem with reward signals updated by IRL procedure and comparing the similarity between trajectories sampled from the policy and demonstrations \cite{russell1998learning}.
%  A brief literature search suggests that the terminology is rather confusing.  Some equal IL with IRL with Apprenticentship leaning, some treat IRL as a special case of IL.  In any case, my finding is that IRL is always associated with demonstration and expert policies instead of being more general than that.  Please take a look at the following links and summarize IRL in a couple of sentences. https://smartlabai.medium.com/a-brief-overview-of-imitation-learning-8a8a75c44a9c; https://en.wikipedia.org/wiki/Apprenticeship_learnin

\vspace{-3pt}\subsection{Our Contributions}\vspace{-3pt}
We propose a new method for RLfD named LEARN-SAM, standing for \emph{Local Ensemble And Reparameterization with Split And Merge of expert policies}. LEARN-SAM parameterizes the policy as a weighted ensemble of the expert-free policy and pre-trained expert policies from demonstration, differentiate the effectiveness of the expert policies in states near vs. far from the states where demonstration has data during the training process, and learns and utilizes the helpful part of an  sub-optimal demonstration to boost the learning speed. The innovation of our work is listed below.
\begin{itemize}
\item The relative weights of the expert-free policy and expert policies  in the ensemble are updated throughout learning, along with the parameters of the former. An expert-free policy receives more (less) weights when it  is responsible for a bad (good) action.
\item We define a new concept, the \emph{$\lambda$-function}, in the policy parameterization to discount the influence of a demonstration in states  relatively far from the demonstrated states. By restricting the effectiveness of expert policies to local regions, the $\lambda$-function helps lower the likelihood of the agent traveling to states with minimal expert instructions and thus improving the learning speed.
\item When the demonstrations are  sub-optimal, they may be instructive in informing the actions on some states but less useful in other states due to either noise or a lack of data on those states. To make the best use of the expert policies trained from them, we propose a \emph{split-and-merge (SAM)} mechanism to split the expert policies by states and then re-group them  to form new expert policies according to how helpful they are in learning acceleration. We theoretically prove the invariant property of the ensemble policy before and after the SAM procedure.
\item We run 6 experiments in the OpenAI gym \cite{brockman2016openai} and Mujoco physics engine \cite{todorov2012mujoco} and compare LEARN-SAM with some of the state-of-the-art RLfD techniques and baseline RL methods. LEARN-SAM achieves faster learning with non-inferior rewards relative to its comparison methods.
\end{itemize}
%Also the weight factors in the linear combination, as parameters, would automatically decrease when expert policy contributes more in decision of bad actions or less in that of good ones than expert-free policy. However lambda function could diminish whole expert policy when part of it are highly misleading, but that does not mean the resting part are not helpful. Therefore we propose split and merge mechanism to re-assemble expert policies, cutting out helpful parts and group them together, so that the weight parameters could find it easier to utilize positive prior knowledge for training process boost. We summarize our contributions as following.

\vspace{-3pt}\subsection{Related Work}\vspace{-2pt}
We review some of the state-of-art research on RLfD and then discuss their differences and connections with our LEARN-SAM procedure.

Some RLfD methods differentiate the effectiveness of expert policies in different states by incorporating regularization terms or including  constraints in the objective function \cite{kang2018policy, jing2020reinforcement, nair2018overcoming}, or by perturbing rewards signals with an additional demonstration-guided shaping term while keeping the optimal policy invariant \cite{brys2015policy, brys2015reinforcement}.  %Other than explicit regularization term in objective function, penalty could be equipped implicitly by perturbing reward signal appropriately. 
Potential-based reward shaping encourages agents to explore demonstrated local areas by granting additional virtual rewards. % computed using metric $d$ similar in our work, when agent come across near demonstrated states. 
However, the effectiveness of reward shaping depends on the specification  of the potential functions given the agent's prior knowledge of the environment. When demonstrations are of low quality, the learning can be very slow. \cite{kang2018policy} proposes the Policy Optimization from Demonstration (POfD) method, inspired by the GAN neural networks (NNs) \cite{goodfellow2014generative}, that uses a parameterized classifier to determine whether a state-action pair is from demonstration and to generate indistinguishable trajectories.

%\subsubsection{Imitation Learning and Reinforcement Learning}
In imitation learning (IL), the agent learns the optimal policy by following or imitating the expert’s decisions  \cite{hussein2017imitation, ho2016generative, ajay2020opal, nair2020awac, singh2020parrot}. There are two main approaches in IL: to pre-train an expert policy from demonstrations via supervised learning via behavior cloning \cite{pomerleau1988alvinn, morales2004learning, torabi2018behavioral} or to use inverse RL to find a cost function under which the expert is uniquely optimal  \cite{russell1998learning, ng2000algorithms, ho2016generative}.% IRL views the demonstrated actions as a sequence of optimal decisions and searches for a reward function under which the policy produces similar trajectories as in demonstrations. 
IL efficiently leverages the prior knowledge from demonstrations to achieve high accumulated rewards at the beginning of the learning process. On the other hand, IL may also get stuck in a local optimum and settle for sub-optimal policies since it  may not explore enough in a large state space with sparse reward signals  in late-stage learning given its assumption that assumes the demonstrations are optimal \cite{jing2020reinforcement}. Work that acknowledges that demonstration data may be noisy exits, such as the the maximum entropy inverse RL paradigm \cite{ziebart2008maximum, wulfmeier2015maximum} that employs a probabilistic approach for inverse RL. 

%which consequent the training process reluctant to sample significant signal from environment. 
%known as Behavior Cloning (BC), treats the action as the target label for each state, and then learns a generalized mapping from states to actions in a supervised manner. Another way, known as Inverse Reinforcement Learning (IRL), views the demonstrated actions as a sequence of decisions, and aims at finding a reward/cost function under which the demonstrated decisions are optimal.
%Imitation could also be accomplished during the training phase by introducing regularization based on discrepancy measures and punishing agents from deviating away from demonstrated trajectory.

%\subsubsection{Hierarchical Reinforcement Learning}

% Similarly, option-based hierarchical RL methods performs RLfD tasks with supervise-learning or regularization techniques \cite{le2018hierarchical, skrynnik2021forgetful}, but in the more general semi-Markov decision process (semi-MDP) framework. Options \cite{sutton1998intra} -- an extension of policy -- activate, maintain, or terminate sub-policies that provide actions achieving sub-goals. Besides the samples on demonstrations, which provide instructions for either sub-policies or options to boost training process \cite{le2018hierarchical}, additional labels are also needed to indicate the sub-goals the agent will accomplish. The labels cane be collected together when demonstrations are sampled or be inferred from other knowledge sources \cite{skrynnik2021hierarchical}

Similarly, option-based hierarchical RL methods performs RLfD tasks via supervise-learning with regularization \cite{le2018hierarchical, skrynnik2021forgetful}, but in the general semi-Markov decision process (SMDP) framework.\footnote{SMDP is considered more general than MDP in that its distributions are conditional on a finite or infinite historical states, actions and reward signals whereas MDP restricts the conditional information to current state. In option-to-option methods\cite{sutton1998intra}, any MDP with a fixed set of options is a SMDP~\cite{puterman2014markov}.} Options \cite{sutton1998intra} -- an extension of policy -- activate, maintain, or terminate sub-policies that provide actions to achieve sub-goals. Besides the samples on demonstrations, which provide instructions to assist training \cite{le2018hierarchical}, additional labels are needed to indicate which sub-goals the agent will accomplish. The labels can be collected together with the demonstrations data \cite{skrynnik2021hierarchical}

% Hierarchical RL generalizes the Markov decision process (MDP) to semi-MDP by promoting policies into option,  a parameterized macro-policy that activates, maintains, or terminates a sub-policy \cite{sutton1999between}. Hierarchical RL often undergoes a significant amount of interactions with the environment and needs help from a large number of parameters for effective use of demonstration and learning \cite{daniel2016probabilistic}. The performance of hierarchical RL is subject to the curse of dimensionality and also depends on the pre-specified number of options \cite{cao2012bayesian}.

Compared to the above-listed work on RLfD where the influence of pre-trained experts on subsequent learning is often fixed, that in in our proposed LEARN-SAM is adaptive to the learning process with the introduction of the $\lambda$ function and the SAM mechanism. In this way, LEAN-SAM effectively acknowledges that the demonstrations may be sub-optimal and thus makes selective usage of demonstrations so to speed up learning rather than following the experts in a blinded way. The $\lambda$ function limits the instructions of the expert policies to regions around their demonstrated states whereas in states where the expert policies have minimal information, the expert-free policy will dominate. In addition, the output of $\lambda$-function varies by state and provides local granularity regarding the effectiveness of an expert policy, as the effectiveness of an expert can vary across the states.  To our knowledge, there is also no existing work similar to the SAM mechanism, which further boosts the efficiency in demonstration usage.

LEARN-SAM, in a broad sense, is still IL, but with more critical and selective usage of the information in demonstrations than most current IL approaches, which becomes clearer when we introduce the method in Section \ref{sec:method} and display the experiment results in Section \ref{sec:experiment}. LEARN-SAM is not strictly inverse RL in that sense that it does search for a reward function from the  demonstrations that could explain the expert behavior, though it employs supervised learning to train an expert policy given demonstration data.  %by splitting an expert policies into different groups and assigning large weights to the groups that are useful for learning.  %The hierarchical RL paradigm could be used to achieve what the $\lambda$ function aims at, but it comes with more computational costs and is a slower training process given the number of parameters involved and the amount of interactions between the agent and the environment. Furthermore,

%----------------------------------------------------------------

\vspace{-4pt}\section{Preliminaries}\vspace{-1pt}
Table \ref{tab:notation} lists the notations and some common definitions in the RL literature as well as in this paper.
\begin{table}[!htb]
\centering\vspace{-6pt}
\begin{tabular}{@{}p{2.5in}|p{3.5in}@{}}
\hline
notation&  name\\
\hline
$\mathcal{S}$&  state space \\
$\mathcal{A}, \mathcal{A}(s)$&   action space (given a state)\\
$T:\mathcal{S}\times\mathcal{A}\times \mathcal{S}\to \R$&    transition function \\
$T(s'|s,a)$&    probability of agent going to state $s'$ from $s$ when taking action $a$ \\
$R(s,a,s')$ with $R: \mathcal{S}\times\mathcal{A}\times\mathcal{S}\to \R$&    reward received by the agent when reaching state $s'$ from $s$ by taking action $a$\\
$\rho_0: \mathcal{S}\to \R$&    probability distribution of the initial state $s_0$ \\
$a_0$&   action taken at the initial state $s_0$  \\
$(\mathcal{S}, \mathcal{A}, T, R,\rho_0, \gamma)$&   discounted MDP (a 6-element tuple)\\
$\gamma \in (0,1]$&   discount factor\\
$\pi:\mathcal{S}\times\mathcal{A}\to \R$&    policy \\
$\pi_{\bs\theta}(a|s)$&    probability of agent taking action $a$ at state $s$, parameterized in $\bs\theta$ (unknown)\\
$V_\pi(s) \coloneqq \E_{a\sim \pi(\cdot|s)}\left[Q_\pi(s,a)\right]$&   state-value function \\
$ Q_{\pi}(s,a)$&   action-value function, where\\
& $\textstyle \E_{\pi}\!\!\left[\!\sum_{t=0}^\infty\!\gamma^t \!R(s_t,a_t,s_{t+1})\bigg| s_0\!\!=\!\!s, a_0\!\!=\!\!a\!\right]$\\
& $\E_\pi=\E_{a_t \sim \pi(\cdot|s_t), s_{t+1}\!\sim\! T(\cdot|s_t,a_t)}$\\
$\mu_\pi(s) =\sum_{t=0}^\infty \gamma^t \mathbb{P}(s_t = s|\pi)$&   discounted visitation frequencies\\
$\rho_\pi(s, a)= \mu_\pi(s)\pi(a|s)$&   occupancy measure\\
$m$&   number of expert policies (given)\\
$D_j$ for $j=1,\ldots, m$&   demonstration set $j$\\
$n_j$&   number of observed demonstrations in set $D_j$\\
$\pi_j$&    expert policy $j$\\
$w_j$&   weight for $\pi_j$ (unknown parameter)\\
$\lambda_j$&   $\lambda$-function of $\pi_j$ (Definition \ref{def:lambda} of Sec \ref{subsec:lambda})\\
$K$&  number of latent experts in  SAM  (Sec~\ref{sec:sam})\\
$\pi_{jk}$&  mini-policies that form expert policy $\pi_j\!=\! \sum_{k=1}^K\!\xi_{jk} \pi_{jk}$\\
$\xi_{jk}$ &   weights of $\pi_{jk}$ in $\pi_j$\\
$\psi_j(s)$&   score function of $\pi_j$ at state $s$\\
\hline
\end{tabular}\vspace{3pt}
\caption{Notations and Definitions} \label{tab:notation}\vspace{-15pt}
\end{table}

\vspace{-4pt}\subsection{Policy Gradient}\vspace{-2pt}
Searching for an optimal policy to maximize the accumulated discounted rewards by a certain time is the goal for the agent in RL.  An optimal policy can be obtained by looping between the value evaluation phase and the policy improvement phase \cite{sutton2018reinforcement}.  Q-learning \cite{watkins1992q} and SARSA \cite{singh1996reinforcement} are two popular algorithms for policy optimization in the value-based paradigm. The policy gradient method is an alternative to the value-based paradigm that measures the quality of the policy with an objective function
\begin{equation}\label{eqn:objective}
\begin{matrix*}[l]
\textstyle\eta(\pi) &  \coloneqq\E_{\pi,s_0\sim \rho_0(\cdot)}\left[\sum_{t=0}^\infty\gamma^t R(s_t,a_t,s_{t+1})\right]\\
& = \E_{s_0\sim \rho_0(\cdot),a_0\sim \pi(\cdot|s_0)}[Q_\pi(s,a)].
\end{matrix*}
\end{equation}
The difference in $\eta$ between  policies $\tilde{\pi}$ and $\pi$ can be calculated from the discounted visitation frequencies $\mu_\pi$ and the advantage  $A_\pi(s,a) = Q_\pi(s,a)-V_\pi(s)$ \cite{kakade2002approximately,schulman2015trust}; that is,
\begin{equation}\label{eqn:policydiff}
\begin{matrix*}[l]
\eta(\tilde{\pi})&  = \eta(\pi) + \E_{\tilde\pi,s_0\sim\rho_0(\cdot)}\left[\sum_{t=0}^\infty \gamma^t A_\pi(s_t,a_t)\right]\\
& = \eta(\pi) + \sum_s \mu_{\tilde{\pi}}(s) \sum_a \tilde{\pi}(a|s)A_\pi(s,a).
\end{matrix*}
\end{equation}
Eqn (\ref{eqn:policydiff}) suggests that $\tilde{\pi}$ will outperform $\pi$ when the second term on the right-hand side is positive. However, this term involves $\mu_{\tilde{\pi}}$, making the direct optimization of Eqn (\ref{eqn:policydiff}) difficult. \cite{schulman2015trust} proposes  a local approximation  $L(\tilde{\pi})$ to $\eta(\tilde{\pi})$ in Eqn (\ref{eqn:policydiff}) for easier optimization. Replacing  $\mu_{\tilde{\pi}}$ in the right-hand side of Eqn (\ref{eqn:policydiff}) with $\mu_\pi$ results in
\begin{equation}\label{eqn:local}
\textstyle L(\tilde{\pi}) = \eta(\pi) + \sum_s \mu_\pi (s)\sum_a \tilde{\pi}(a|s)A_\pi(s,a).
\end{equation}
For policy $\pi_{\bs\theta}$ paramaterized in ${\bs\theta}$ that is  first-order differentiable $\forall \bs\theta\!\in\!\Theta$,  $L(\pi_{\tilde{\bs\theta}})$ is a first-order approximation to  $\eta(\pi_{\tilde{\bs\theta}})$  in the sense that 
\begin{equation}\label{eqn:L}
\ds L(\pi_{\tilde{\bs\theta}}) \ds = \eta(\pi_{\tilde{\bs\theta}}) \mbox{ and }
\ds\nabla_{\bs\theta} L(\pi_\theta)|_{\bs\theta=\tilde{\bs\theta}} \ds = \nabla_{\bs\theta} \eta(\pi_{\bs{\theta}})|_{\bs\theta=\tilde{\bs\theta}}
\end{equation}
for  $\forall \bs\theta\!\in\!\Theta$ and  $\forall \tilde{\bs\theta}\!\in\!\Theta$, where $\tilde{\bs\theta}$ is the parameter associated with policy $\tilde{\pi}_{\bs\theta}$. 
Eqn (\ref{eqn:L}) implies that a sufficient small step that improves $L$ also improves $\eta$. Trusted Region Policy Optimization (TRPO) \cite{schulman2015trust} is a popular algorithm that maximizes $L_{\pi_{\tilde{\bs\theta}}}(\pi_{\bs\theta})$ with a trust region constraint  to achieve policy improvement for a given $\delta$:
\begin{equation}\label{eqn:TRPO}
\ds\max_{\bs\theta} L_{\pi_{\tilde{\bs\theta}}}(\pi_{\bs\theta})
 \text{ s.t. } \max_{s\in\mathcal{S}}\mathcal{D}_{KL}(\pi_{\tilde{\bs\theta}}(\cdot|s)\|\pi_{\bs\theta}(\cdot|s)) \leq \delta.
\end{equation}
Our RLfD method employs the policy-gradient framework. 

\vspace{-1pt}\subsection{Reinforcement Learning from Demonstration}\vspace{-2pt}
Demonstrations are often sampled from policies executed by a human expert or a well-trained policy from previous RL tasks. In the policy-gradient framework, RLfD searches for an optimal policy $\pi$ that maximizes $\eta(\pi)$ given demonstration $D$, where $D$ is a collection of a set of trajectories of state-action pairs in chronological order. Let $\rho_\pi$ and $\rho_E$ be the occupancy measures associated with the policy to be optimized and an expert policy, and $\mathcal{D}(\cdot \| \cdot)$ be a discrepancy measure between two occupancy measures \cite{jing2020reinforcement}.  One of the RLfD methods is to  incorporate $\mathcal{D}(\rho_\pi \| \rho_E)$ in the objective function to encourage the agent to take similar actions and to reach similar states as the expert during learning, such as 
\begin{equation}\label{eqn:loss_rlfd}\vspace{-3pt}
\min_\pi L_\pi = -\eta(\pi) + \lambda \cdot \mathcal{D}(\rho_\pi \| \rho_E),
\end{equation}
where the discrepancy measure $\mathcal{D}(\rho_\pi \| \rho_E)$ is a regularization term that promotes the similarity between $\rho_\pi$ and $\rho_E$. %Another approach is to reduce $\rho_\pi$ as a one-step occupancy measure, that is $\mathbb{P}(s'|\pi, s)$. 

\cite{kim2013maximum} proposes the maximum mean discrepancy-IL (MMD-IL) framework. Given $\mathcal{F}$ as a class of functions $f: \mathcal{S}\to \R$, MMD between $\rho_\pi$ and $\rho_E$ could be estimated using demonstrations $D_1$ and $D_2$ sampled from them; that is,
\begin{equation}\label{eqn:MMD}
\text{MMD}[\mathcal{F},D_1, D_2] = \sup_{f\in\mathcal{F}}\bigg(\frac{1}{m}\sum_{i=1}^m f(s_i) - \frac{1}{n}\sum_{j=1}^n f(s_j)\bigg)
\end{equation}
where $s_i$, $s_j$ are the states from $D_1$ and $D_2$  and $m$ and $n$ are the number of state-action pairs in $D_1$ and $D_2$, respectively. $m$ can be as small as 1, in which case, MMD measures the dissimilarity between a single state and a set of states.  At each iteration of the MMD-IL procedure, the discrepancy between the current state to the demonstration states is evaluated, which can be used to determine the source of the next action -- from the expert policy, from a vanilla RL training procedure running at the same time, or from an online oracle expert \cite{kim2013maximum}.

We consider sub-optimal demonstrations in this study, where the sub-optimality may be attributed to two sources: data quality and data quantity. Before describing sub-optimal demonstrations, it helps to understand what an optimal expert and its demonstration are.  
\begin{definition}[Optimal Expert Policy~\cite{jing2020reinforcement}]\label{def:perfect}
Given loss function $L_\pi$, an optimal expert policy $\pi_{\theta+}$ is defined as
\begin{equation}
    \pi_{\theta+} \in \bigg\{\pi: {\arg\min}_{\pi} L_\pi \text{ and } \frac{\partial L_\pi}{\partial \theta} = 0\bigg\}
\end{equation}
\end{definition}
optimal demonstrations are samples from the optimal expert policy  in Definition~\ref{def:perfect} and  achieve optimized aggregated reward. By contrast, a sub-optimal demonstration does not satisfy the condition for optimality;  it could be a sample from a sub-optimal policy that does not achieve optimized aggregated reward (lower quality), a sample from the optimal policy but the data is noisy/contaminated (lower quality), or it does not contain insufficient state-action pairs (data sparsity). %With sub-optimal demonstrations, the discrepancy  between the expert policy and the parameterized policy (the second term of Eqn~\eqref{eqn:loss_rlfd} may be biased, so is the subsequent gradient calculations during the policy update~\cite{jing2020reinforcement}.

%------------------------------
\section{Methodology}\label{sec:method}

With demonstrations $D_j = \{(s_{jt}, a_{jt})\}_{t=1}^{n_j}$ for $j=1,\ldots,m$, the corresponding expert policies $\pi_j$ can be pre-trained in a  supervised manner without interacting with the environment \cite{ramirez2021model}. The learned expert policies $\{\pi_j\}_{j=1}^m$ can then be combined with the expert-free policy $\pi_{\bs\theta}$, to obtain an ensemble policy $\pi_{\bs\zeta}$ 
\begin{equation}\label{eqn:ensemble}
\textstyle \pi_{\bs\zeta}(a|s)=(1-\sum_{j=1}^mw_j)\pi_{\bs\theta}(a|s)+\sum_{j=1}^mw_j\pi_j(a|s),
\end{equation}
where $\bs\zeta=\{\bs\theta,\w\}$ and  $\w=\{w_j\}_{j=1}^m$ contains the unknown weight associated with each expert policy that is to be estimated during the RL process. Policy-gradients methods such as TRPO in Eqn \eqref{eqn:TRPO} can be used to estimate $\bs\zeta$. \footnote{\cite{fu2019automatic, khurana2018ensembles, gao2018reinforcement, wu2019imitation} weight the information in demonstrations by optimizing a weighted linear combination of cost functions rather than through policy ensemble as in Eqn \eqref{eqn:ensemble}. We conjecture that the two formulations lead to similar performance.}

Albeit an intuitive approach, learning of the ensemble formulation in Eqn (\ref{eqn:ensemble}) can be slow given the  sub-optimality of the demonstrations. If the demonstrations contain sub-optimal or even  false action information, the ensemble policy that utilizes the demonstrations may get stuck in a local optimum and never get a chance to get out in a reasonable or realistic time frame.  In addition,  the recorded state-action pairs from demonstrations are formulated as Markov chains, a one-dimensional manifold embedded in a potentially high-dimensional space that may not always capture the information of the whole state space.

We make two major improvements over the ensemble policy formulation in Eqn (\ref{eqn:ensemble}) with a new parameterized policy scheme for efficient RL in the presence of  sub-optimal demonstration, with the introduction of a new concept the $\lambda$ function, and the development of the SAM mechanism. We name the new procedure \emph{Local Ensemble And Reparameterization with Split And Merge} of expert policies (LEARN-SAM). We will cover both elements below, along with an algorithm to implement the LEARN-SAM procedure.

\subsection{Policy Ensemble with \texorpdfstring{$\lambda$-Function}{}}\label{subsec:lambda}\vspace{-3pt}
The $\lambda$-function  controls the influence of an expert policy on state $s$ with an reformulated  policy ensemble as
\begin{equation}\label{eqn:ensemblelambda}
\!\!\pi_{\bs\zeta}(a|s)\!=\!\bigg(\!1\!-\!\sum_{j=1}^m\! \lambda_j(s) w_j\!\!\bigg)\pi_{\bs\theta}(a|s)\!+\!\sum_{j=1}^m \!\lambda_j(s) w_j\pi_j(a|s),\!
\end{equation}
where  $w_j \in (0, 1)$, $\sum_{j=1}^m w_j \leq 1$, and  $\lambda_j(s)\in[0,1]$ $\forall\;s\in\mathcal{S}$. The intuition behind the $\lambda$-function concept is that the guidance expert policy $\pi_j$ cannot be generalized to states far away from those in demonstrations. Therefore, one would down-weight the expert's guidance on a state where the guidance is not good or when there is a lack of guidance (no data), instead of blindly applying the same global weight $w_j$ to the expert policy for all the states. In contrast,  $\lambda_j(s)$ is a local weight, varies by state and provides the needed local granularity for $\pi_j$.  The more confident we are about the expert guidance on a state $s$, the larger $\lambda(s)$ is. There  are potentially many different formulations of the $\lambda$-function; Definition \ref{def:lambda} provides one such formulation, which is used in the experiments in Section \ref{sec:experiment}.
\begin{definition}[the $\lambda$-function]\label{def:lambda}
Given a set of demonstrated states $D_{n\times 1} \subset \mathcal{S}$ and a metric function $d:\mathcal{S}^{n+1}\to \R^+$ that measures the distance between a state $s$ to $D$, we define $\lambda(s) = \exp(-\varphi(d(s, D)))$ where $\varphi: \R^+ \to \R^+$ is a monotonically increasing bijection.
\end{definition}
It is obvious that $\lambda(s)$ monotonically decreases with $d$, $\varphi$ defines the decreasing rate of $\lambda$ in $d$, and $\lambda(s) = \exp(-\varphi(0))=1$. It is also desirable to have the properties 1) that $d$ is closed in the sense it is the same type of $d$ after reparameterization via the SAM procedure (see Sec \ref{sec:sam}), and 2) when $\forall s \in D$, $d(s, D) = 0$; that is, the $\lambda$ function fully empowers the expert policy when $s$ belongs to the demonstration set $D$. %$\lambda$ should not be larger than $1$ nor smaller than $0$ because it might lead to weights $\w_j$ larger than $1$ or negative, hence the whole policy~\ref{eqn:ensemblelambda} is not well-defined.
%\footnote{We do not recommend using MMD as $d$. The reason has to do with the invariant property associated with the SAM mechanism presented in Section \ref{sec:sam} and is detailed in the appendix.}
% Other metrics such as the MMD given in Eqn \eqref{eqn:MMD} \cite{kim2013maximum}, the Wasserstein distance \cite{dadashi2020primal}, the f-Divergence \cite{ke2020imitation}, and the integral probability metrics \cite{muller1997integral} can also used. Although Some of these metrics are distribution based, they still can be approximated with demonstrations and trajectories sampled in RL procedure. 

The metric function $d$ can be any legitimate distance measure  when the state space is continuous, such as the Euclidean distance from a point to a set. For discrete state spaces, with reasonable topological or geometrical assumption can be made, $d$ can still be defined. For example, if the states are nodes of a graph, the shortest path length between two states is a natural choice for $d$; if the states are blocks in a grid world \cite{sutton2018reinforcement}, the Manhattan distance can be used as $d$. When there is reasonable topological or geometrical assumption, 0-1 distance could be used for d, meaning the agent would only consider executing a demonstrated actions when it is in the demonstrated state precisely. 

There are also various choices for $\varphi$, e.g., a square function or a linear function with the latter yielding larger values of $\lambda$ than the former; in other words, $\lambda$ decays at a slower rate  with the linear function as $s$ gets further away from $D$ so the expert policy can be evaluated more thoroughly by the agent. %closed is not a requirement though nice to have
%, $\lambda(s)$ becomes a Gaussian kernel. If  $\varphi$ is an linear function with slope $h$ and no intercept,  which is what we used in the experiments, $\lambda(s) = \exp(-h\cdot \min_{s'\in D}\|s-s'\|)$ is the maximum Laplacian kernel between $s$ and $D$. Compared to the  Gaussian kernel, the Laplacian kernel yields more large values of $\lambda$. In other words, $\lambda$ decays at a slower rate as $s$ gets further away from $D$ so the expert policy can be evaluated more thoroughly by the agent. $d$ is the Euclidean distance and

\vspace{-3pt}
\subsection{Reparameterization of Expert Policies via Split and Merge}\label{sec:sam}\vspace{-3pt}
The second improvement we make to the ensemble policy formulation in Eqn (\ref{eqn:ensemble}) is the introduction of the Split and Merge (SAM) mechanism. SAM evaluates an  expert policy state by state (mini-policies), then groups the mini-policies into different categories according to some criterion on helpfulness in learning (the Split step), and finally combines the mini-policies in the same category across all the expert policies to form a new expert policy (the Merge step).

The intuition behind the SAM mechanism is as follows. The $\lambda$ function restricts the influence of expert policies to local regions where there is demonstration data; however, it cannot tell a good action from a bad one and thus does not have the power to prevent inappropriate actions from being included in the ensemble policy. In other words, bad actions can still be taken, slowing down learning or leading the agent to a local optimum. There is a need to introduce a mechanism to measure the quality of an expert state by state, and down-weight its influence if its action in a state is not helpful for learning. Toward that end, we introduce the SAM mechanism.

Specifically, we assume each of the $m$ expert policies $\{\pi_j\}_{j=1}^{m}$ in each state $s$ can be represented as a mixture of $K\ge2$ policies  $\{\pi_k\}_{k=1}^K$ ((or notation simplicity, we use $\pi_j$ and $\pi_{jk}$ to represent $\pi_j(s|a)$ and $\pi_{jk}(s|a)$, respect9vely), where the $K$ policies are different in terms of the optimality of the state-action value at state $s$.
\begin{align}\label{eqn:mixture}\vspace{-3pt}
&\pi_j\!=\!\textstyle\sum_{k=1}^K\!\xi_{jk}(s) \pi_{jk},\\
&\mbox{where $\xi_{jk}(s)\!\in\![0, 1]$ and $\sum_{k=1}^K\!\xi_{jk} (s)\!= 1$}.\notag
\end{align}
The linear decomposition of the expert policy holds automatically if we let $\pi_{jk}\equiv\pi_j$ for $k=1,\ldots,K$, a convenient choice for the reparameterization and SAM mechanism; other types of formulation of $\pi_{jk}$ can also be used as long as the linear decomposition still holds. 

In terms of choice of $K$, any concept that makes sense for a RL task can be used. An intuitive criterion is helpfulness or effectiveness in  learning assistance. We may use ordinal categories such as  very unhelpful to unhelpful to neutral to helpful to very helpful ($K=5$); or as simple as two categories -- helpful vs.~unhelpful ($K=2$). In the case of $K=2$, when $\pi_j$ provide good guidance at state $s$, the weight $\xi_{j1}(s)$  of the helpful component $\pi_{j1}$ would be larger than the weight $\xi_{j2}(s)$ of the unhelpful component $\pi_{j2}$. We use $K=2$ in all the experiments in Section \ref{sec:experiment}. %which seems to be sufficient to demonstrate the effectiveness of SAM
$K>2$ can be used, leading to more granularity in identifying policies of different optimality and potentially better performance of of SAM, but at a larger computational cost. In practice, $K$ can be chosen via a grid search with an appropriate evaluation metric (e.g., aggregated reward with a fixed number of epochs); $K$ can also set be set at a relatively large value if there is no sufficient data to inform the granular separation. If the actual $K$ turns out to be smaller than the pre-set $K$, the weights of some of the components should be around 0 by definition.

After the reparameterization, the ensemble exert policy can be re-expressed in terms of the \emph{latent experts} $\{\pi'_k\}_{k=1}^K$. For example, when $K=2$, there are two latent experts, representing the helpful and unhelpful experts, respectively. The helpful latent expert will be up-weighted as a whole and participate in the subsequent learning; and the unhelpful latent expert will still participate in the subsequent learning with a smaller weight than the helpful latent expert.\footnote{Mathematically, the weight for the unhelpful expert cannot be 0 to the sake of the invariant property. In practical implementation, its weight will get smaller and diminishes toward 0 after each SAM application. When its weight is close to 0, we can throw away the  non-helpful experts to further improve the learning efficiency with minimal impact on the invariant property.} We now establish the invariant property of the ensembled expert policy with reparameterization in the Proposition \ref{prop:splitmerge}, which provides theoretical guarantees for the convergence of the ensemble policy optimization.
\begin{proposition}\label{prop:splitmerge}
Let \vspace{-3pt}
\begin{align}
\lambda'_k(s)& \textstyle = \max_j\{\lambda_j(s): w_j \xi_{jk}(s) > 0\},\label{eqn:lambda_k}\\
\beta_{jk}&\textstyle = w_j\lambda_j\xi_{jk}(s)/\sum_{j=1}^m w_j \lambda_j \xi_{jk}(s),\mbox{ and }\label{eqn:beta_jk}\\
\pi'_k&\textstyle = \sum_{j=1}^m \beta_{jk}\pi_{jk}\mbox{ be the $k$-th latent expert policy}
\end{align}
for $k=1,\ldots,K$. There exist $w'_k \in (0, 1)$ such that, for any given $s$, $\sum_{k=1}^K w'_k \leq 1$ and 
\begin{equation}\label{eqn:repara}
\textstyle\sum_{j=1}^m w_j \lambda_j(s) \pi_j = \sum_{k=1}^K w'_k \lambda'_k(s) \pi'_k.
\end{equation}
\end{proposition}
\begin{proof}
$\lambda'_k(s) = \max_j \{\lambda_j(s): w_j\xi_{jk}(s) > 0\}$ is closed with the reparameterization as
\begin{align}
    \lambda'_k(s) & = \max_{j}\{\exp(-\varphi(d(\cdot, S_j)): w_j \xi_{jk}(s) > 0\} \notag\\
    & =\exp(-\varphi(\min_{j}\{d(\cdot, S_j): w_j \xi_{jk}(s) > 0\}))\notag\\
    & = \exp(-\varphi(d_k(\cdot, S_k))),\notag
\end{align}
where $d_k \coloneqq \min_{j}\{d(\cdot, S_j): w_j \xi_{jk}(s) > 0\}$ and $S_k \coloneqq \bigcup_{j=1}^m \{S_j: w_j \xi_{jk}(s) > 0\}$. When $s \in S_k$ and $\forall k$, a single element in $\{\lambda_j:w_j\xi_{jk}(s) > 0\}$ is 1, also implying $\lambda'_k(s)= 1$ in this case.  Since $\pi_j\!=\!\sum_{k=1}^K \xi_{jk}(s) \pi_{jk}$, 
\begin{equation}
    \sum_{j=1}^m w_j \lambda_j \pi_j\!=\!\sum_{j=1}^m w_j \lambda_j \sum_{k=1}^K \xi_{jk}(s)\pi_{jk}\!
    =\!\sum_{k=1}^K \sum_{j=1}^m w_j \lambda_j \xi_{jk}(s) \pi_{jk} \!=\! \sum_{k=1}^K \alpha_k \sum_{j=1}^m \beta_{jk}\pi_{jk},\notag
\end{equation}
where $\alpha_k\! = \!\sum_{j=1}^m w_j \lambda_j \xi_{jk}(s)$ and $\beta_{jk}\!=\!  w_j\lambda_j\xi_{jk}(s)/\alpha_k \geq 0$. Since $\sum_{j=1}^m \beta_{jk}\! =\! 1$  for each $k=1,\ldots,K$, the latent expert policy $\pi'_k\!=\!\sum_{j=1}^m \beta_{jk}\pi_{jk}$ is well-defined. Eqn \eqref{eqn:repara} is obtained by letting $w'_k = \alpha_k / \lambda_k'$. %Besides, the metric $d_k(s, S_k)$ is well-defined and induce $\lambda'_k$ as shown above.
\vspace{-12pt}
\end{proof}
Proposition \ref{prop:splitmerge} guarantees that the expert ensemble in Eqn \eqref{eqn:ensemblelambda} stays the same  before and after the reparameterization (i.e.,$\{w_j,\lambda_j, \pi_j\}_{j=1,\ldots,m}$ vs. $\{w'_k, \lambda'_k,\pi'_k\}_{k=1,\ldots,K}$) during the optimization. 
The SAM mechanism can be applied at each epoch during learning or less frequently given a pre-specified schedule.  The former is computationally more costly as  $w'_k$ needs to be estimated after each application of the SAM procedure, but too few applications would not make the best use of the SAM mechanism, at least during the early learning stage. As the learning continues, we expect the $K$ latent expert policies to stabilize and that the SAM mechanism can be applied less frequently compared to the early stage.
%If an agent is familiar with the environment and has some prior knowledge on which states might benefit more from the usage of the SAM procedure, where to apply the SAM mechanisms can be pre-specified before learning.  In the experiments in Sec \ref{sec:experiment}, without prior knowledge,  we apply the SAM mechanism at every step. 

\vspace{-5pt}\subsection{Choices of \texorpdfstring{$\pi_{jk}$}{} and \texorpdfstring{$\xi_{jk}$}{} in SAM}\vspace{-3pt}
One immediate and convenient option for $\pi_{jk}$ is to let $\pi_{jk}=\pi_j$ for $k=1,\ldots,K$, in which case, $\xi_{jk}(s)$ can then be regarded as the proportion of $\pi_j$ allocated to category $k$.  Other choices of $\pi_{jk}$ can be used. For example, if $\pi_{jk}$ is a Gaussian distribution, then $\pi_j=\sum_{k=1}^K \xi_{jk}(s) \pi_{jk}$ is formulated as a mixture Gaussian distribution. For $\xi_{jk}(s)$, we may define a scoring function  to assign weight $\xi_{jk}(s)$ to different categories $k=1,\ldots,K$ with the constraint $\sum_{k=1}^K \xi_{jk}(s) = 1$  $\forall j$ given a state $s$.  
$\psi_j(s)$ measures the contribution of expert policy $\pi_j$ to learning acceleration in state $s$. Definition \ref{def:score} provides the formal definition on the scoring function.
\begin{definition}\label{def:score}
Given the acting policy $\pi$ and an expert policy $\pi_j$ for $j=1,\ldots,m$, the scoring function $\psi_j(s): \mathcal{S} \to  \R$ for $\forall s \in \mathcal{S}$ is
\begin{equation}\label{eqn:psi}
\psi_j(s) = \sum_{a\in \mathcal{A}(s)} \frac{A_\pi(s, a)}{|V_\pi(s)|}\pi_j(a|s)
\end{equation}
\end{definition}\vspace{-3pt}
%Acting policy indicates the policy of agent is following. off-policy is the policy that agent is not performing, but we are evaluating the value of this policy, even though we are performing another policy. value= (expected future aggregated discounted rewards) 
In our setting, the acting policy is the ensemble policy in Eqn \eqref{eqn:ensemblelambda}. The advantage $A_\pi(s,a)\!=\! Q_\pi(s,a)\!-\! V_\pi(s)$ measures how much more  value an action $a$ achieves in state $s$ compared with the average value of all potential actions with the ensemble policy $\pi$. If a positive (negative) advantage is detected, then $a$ can be said to benefit (harm) learning given policy $\pi$ and state $s$.  $A_\pi(s,a)\pi_j(a|s)$ quantifies the contribution of expert $j$ to the advantage of $a$ in state $s$. $V_\pi(s)$ is a scale parameter  so that $\psi_j(s)$ is within the same order of magnitude across states for the convenience of using the same function to calculate $\xi_{jk}(s)$ \footnote{The magnitude of the value function tends to be larger for states further away from the terminal state.}. All taken together, $\psi_j(s)$ measures the overall contribution of expert policy $\pi_j$ in state $s$, all actions considered.
%The scaling via $V_\pi(s)$ is not necessary as it can be incorporated in later steps of calculating $\xi_{jk}(s)$ given $\psi_j(s)$
%The decision of taking $a$ is made by multiple expert policies and expert-free policy, so intuitively the score function $\psi_j$ should be defined as the contribution from each expert policy $\pi_j$.

There are multiple ways to construct $\xi_{jk}(s)$ from $\psi_j(s)$ (referred to as the \emph{grouping function} $g$) and we suggest two approaches below. The first is the softmax function. Let $b_k \in \R$ be the oracle score in class $k$, which monotonically increases with $k$. For a given $c \in (0, +\infty)$ and $\{b_k\}_{k=1}^K$,
\begin{equation}\label{eqn:psi-softmax}\vspace{-3pt}
\xi_{jk}(s) = \frac{\exp(-c|\psi_j(s) - b_k|)}{\textstyle\sum_{k'=1}^K \exp(-c|\psi_j(s) - b_{k'}|)}.
\end{equation}
For the second formulation for $g$, we use the cumulative density function (CDF) of normal distribution as in\vspace{-3pt}
\begin{equation}\label{eqn:psi-probit}\vspace{-3pt}
\xi_{jk}(s) = \Phi(b_{k}; \psi_j(s), c^{-1}) - \Phi(b_{k-1};\psi_j(s), c^{-1}),
\end{equation}
where $\Phi(\cdot;\psi_j(s), c)$ is the CDF of $\mathcal{N}(\psi_j(s),c^{-2})$, and $b_k$ is the cutoff between categories $k-1$ and $k$ with $b_0 = -\infty$ and $b_K = +\infty$.  $c$ and $\mathbf{b}$ are hyper-parameters in Eqns \eqref{eqn:psi-softmax} and  \eqref{eqn:psi-probit}. The larger $c$ is, the more separate the $K$ categories are.

%The classifier for SAM is not necessarily directly computed from score. For example, given total number of classes $K$, we could built a fully connected neural network taking state as input and $K$-dimensional weights as output by using Softmax as the activation function for output layer. The labels for training could be the binary measure based on score $\psi_j$ total reward collected. However in this way we lost the interpretation for each class. 

%--------------------------------------------
\vspace{-3pt}\subsection{Implementation of LEARN-SAM}
The LEARN-SAM procedure requires the specification or tuning of the following hyperparameters: those associated with $\xi_{jk}$ (e.g, $c$, $\{b_k\}_{k=0}^K$ in Eqns \eqref{eqn:psi-softmax} and \eqref{eqn:psi-probit}), a discrepancy measure $d$, and bijection $\varphi$ in the $\lambda$-function. The parameters to be estimated are  $\bs\zeta=\{\bs\theta,\bs w\}$ in Eqn \eqref{eqn:ensemblelambda}. We use the same loss function as TRPO given in Eqn \eqref{eqn:TRPO} with $\pi_{\bs\theta}$ replaced by the ensemble function $\pi_{\bs\zeta}$ specified in Eqn \eqref{eqn:ensemblelambda} (users often need to specify hyperparameter $\delta$).

Every time the SAM mechanism is applied, $\pi_{\bs\zeta}(a|s)$ in Eqn \eqref{eqn:ensemblelambda} is reparameterized through Eqn \eqref{eqn:repara}. Set $\pi_{jk}\! = \!\pi_j$ for $k\!\!=\!\!1,\ldots,K$; %For notation simplicity, we use the first application of the SAM mechanism in a RL task to illustrate the reparameterization of the policy (the rest of the applications are the same except for the change in the dimensionality of some vectors and matrices). 
denote the expert policies before and after the application of SAM by $\bs\pi\!=\![\pi_j]_{m\times 1}$ and $\bs \pi'\!=\! [\pi_k]_{K\times 1}$, respectively; let $\bs w\!=\![w_j]_{m\times 1}$ and $\bs w'\!= \![w_k]_{K\times 1}$ be the corresponding weights, $\Xi(s)\! = \![\xi_{jk}(s)]_{m\times K}$ be the transition matrix between $\bs\pi$ and $\bs\pi'$, and $M_{K\times m}(s)$ be a matrix containing a single 1 per column, the position of which in column $j$ ($j\!=\!1,\ldots,m$) corresponds to the index of maximum $\lambda_j(s)$ where $w_j\xi_{jk}(s)\!>\!0$. The reparameterization  can be written as\footnote{For notation simplicity, we drop $s$ in $M(s),\Xi(s), \bs \lambda(s)$ and $\bs \lambda'(s)$.}\vspace{-3pt}
\begin{align}
\label{eqn:linear_lambda} \bs \lambda' &  = M \bs\lambda \\
\label{eqn:linear_weight}\bs w' &= (\Xi^T \odot ((\bs 1_{K\times 1} \oslash \bs\lambda')\bs\lambda^T)) \bs w\\
\label{eqn:linear_policy}\bs \pi'\!&=([(\bs w \odot \bs \lambda)(\bs 1_{1\times K} \oslash ((\bs w\odot \bs \lambda)^T\Xi))] \odot \Xi)^T \bs \pi,\vspace{-3pt}
\end{align}
where $\odot$ and $\oslash$ refer to the element-wise matrix multiplication and division; $\bs 1$ is a row vector of 1's; and  $\Xi$ can be calculated from Eqns \eqref{eqn:psi-softmax} and \eqref{eqn:psi-probit} given score $\bs\psi = [\psi_j]_{m\times1}$. Eqns \eqref{eqn:linear_lambda} to \eqref{eqn:linear_policy} implies that the reparameterization from $(\bs\lambda, \bs w, \bs\pi)$ to $(\bs\lambda', \bs w', \bs\pi')$ is a linear transformation for each of the 3 elements, which remains linear after an arbitrary number of reparamterization rounds.

Algorithm \ref{alg:sm} lists the steps for implementing LEARN-SAM. The convergence of the algorithm can be evaluated by eyeballing the trace plots of and monitoring the numerical changes  in the accumulated exact rewards, the objective function, or a composite function of the parameters $\boldsymbol{\eta}$ over epochs.  If the trace plots reach some plateau or the change between two adjacent epochs is below a pre-specified threshold, the algorithm convergences and can stop.
\begin{algorithm}[H]
\caption{The LEARN-SAM procedure}\label{alg:sm}
\begin{algorithmic}[1]
\REQUIRE{Environment $(\mathcal{S}, \mathcal{A}, T, R,\rho_0, \gamma)$, demonstrations $\{D_j\}_{j=1}^m$, state-distance metric $d$, hyperparameters for defining $\xi_{jk}$ ($c$ and $b_k$)} ($\delta$ in Eqn \eqref{eqn:TRPO} if TPRO is used as the optimization procedure).
\STATE Train expert policy $\pi_j$ from $D_j$ for $j=1,\ldots,m$ via NNs (other supervised learning procedures can be used).
\STATE Initialize a NN for expert-free policy $\pi_{\bs\theta}$ and a NN for value $V$.
\STATE Calculate the $\bs\lambda=\{\lambda_j\}_{j=1}^m$ function in Definition \ref{def:lambda}.
\STATE Initialize the ensemble policy $\pi_{\bs\zeta}$ in Eqn \eqref{eqn:ensemblelambda}.
%\STATE convergence $\leftarrow0$
%\WHILE{convergence $=0$}
\FOR{epoch $l=1,2,\cdots$}
\STATE Collect data $\{(s_0, a_0, r_0), (s_1, a_1, r_1), \cdots\}$.
\STATE Update $V(s)$, $A(a, s)$ and compute $\bs\psi$ given the collected data.
\STATE Evaluate $\bs\Xi$  via Eqn~\eqref{eqn:psi-softmax} given $\bs\psi$.
\STATE Calculate $\bs\lambda'$ via Eqn~\eqref{eqn:linear_lambda}.
\STATE Calculate $\bs w'$ via Eqn~\eqref{eqn:linear_weight}.
\STATE Update policy $\bs\pi'$ via Eqn~\eqref{eqn:linear_policy}.
%\STATE Compute the weight $1\!-\!\|\bs w'\!\odot\!\bs\lambda'\|_1$ for expert-free-policy $\pi_{\bs\theta}$. % as sum_k (w'_k lambda'_k) = sum_j(w_j lambda_j)
\STATE Update $\bs\zeta' = \{\bs\theta, \bs w'\}$ using TRPO (Eqn~\eqref{eqn:TRPO}) or other policy gradient methods
\STATE $\bs\zeta \leftarrow \bs\zeta'$, $\bs\lambda \leftarrow \bs\lambda'$, $\bs \pi \leftarrow \bs\pi'$
\ENDFOR
%\STATE convergence $\leftarrow1$
%\ENDWHILE
\RETURN Policy $\pi_{\bs\zeta}$
\end{algorithmic}
\end{algorithm}

\begin{figure}[!htb]
\centering\vspace{-12pt}
\includegraphics[width=0.8\textwidth]{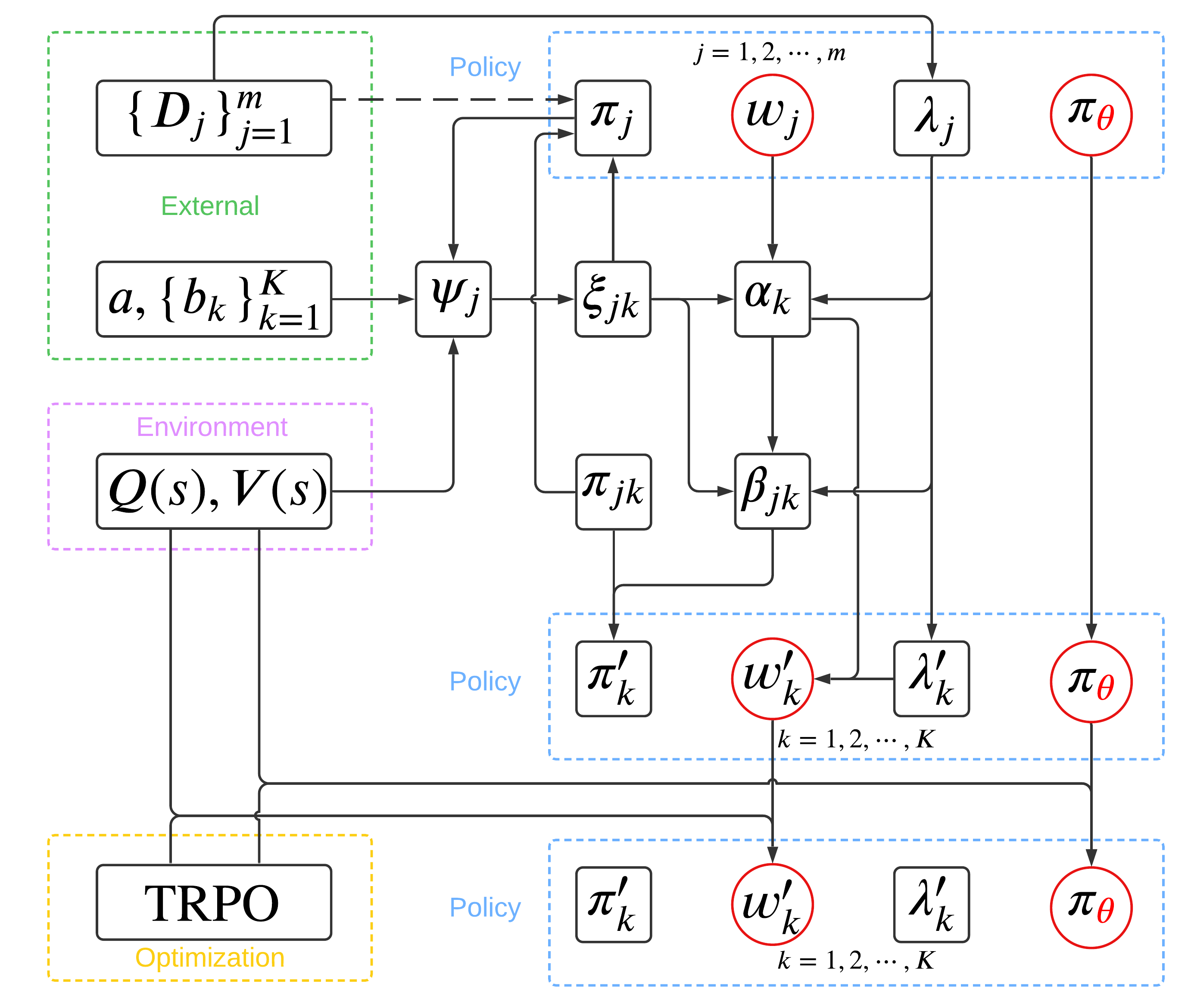}\vspace{-6pt}
\caption{The LEARN-SAM procedure for RLfD}
\label{fig:flow}\vspace{-6pt}
\end{figure}

We summarize the LEARN-SAM procedure in Fig.~\ref{fig:flow}. The
red circles represent the unknown parameters to be estimated by LEARN-SAM ($\w_j,\bs\theta$),  the nodes within the green-dashed-bordered rectangle represent either the hyperparameters that users need to specify (e.g, $c, \mathbf{b}$ for the grouping function $g$ for $\xi_{jk}$ calculation) or the observed data (e.g., demonstrations $D_j$), and the black rectangles represent the parameters or functions specified by users ($\pi_j, \pi_{ik}$) or calculated during the LEARN-SAM procedure (e.g, $\psi,\alpha_k$, etc).  The arrow from demonstration $\{D_j\}_{j=1,\ldots,m}$ to expert policies $\pi_j$ is in 
dash line, meaning that the expert policies $\{\pi_j\}_{j=1,\ldots,m}$ are estimated once before the LEARN-SAM procedure rather than being estimated at every epoch.  The diagram shows the reparameterization and optimization in a single epoch. At the end of the epoch, the new state-action trajectory data are collected and parameters $\{w_j\}_{j=1,\ldots,m}$ and $\bs\theta$ are updated and a new epoch starts with the newly updated $w_j\leftarrow w'_k$, $\pi_j\leftarrow \pi'_k$, and the newly-calculated $\lambda_j\leftarrow \lambda'_k$.

%------------------------------
\section{Experiments}\label{sec:experiment}
We compare LEARN-SAM against several state-of-the-art methods for RLfD  and some baseline approaches in six continuous robotic controlling simulation environments implemented in OpenAI gym and Mujoco (Reacher-v2, walker2d-v3, HalfCheetah-v3, Ant-v3, Hummanoid-v3, DoublePendulum-v2)  \cite{duan2016benchmarking, brockman2016openai, todorov2012mujoco}. In each experiment, we firstly train an expert policy with TRPO then use it to generate a single  sub-optimal demonstration with Gaussian noise added to each action. The policies and value function are parameterized via fully connected NNs with two hidden layers consisting of 64 nodes per layer and with the $\tanh$ activation function. %in all the methods for comparability. 
The performances of the methods are compared after the agent interacts with the environment for the same number of times. We run 10 repeats per experiment in the same experiment settings but with different random seeds. 
% the randomness may come various sources, initial conditions, optimization, and sampling during training, etc. 

The reward set-ups in the 6 experiments belong to one of the following 3 types of reward sparsification.
\begin{itemize}[leftmargin=9pt]
\item Type 1: Agent receives $+1$ reward when it reaches a terminal state; no reward granted otherwise (Reacher-v2);
\item Type 2: Agent receives $+1$ reward when it survives a crash and moves in any direction  (walker2d-v3, HalfCheetah-v3, Ant-v3, Hummanoid-v3); %the agent in the environment, like an ant in Ant-v3, human-like robot in Humanoid-v3, can die if it encounters a large crash
\item Type 3: Agent receives $+1$ reward when the pole reaches higher than 0.89; this is specifically for the InvertedDoublePendulum task (DoublePendulum-v2).
\end{itemize}
%For each experiment, we firstly train expert policy (indicated as Expert in Table~\ref{tab:result}) with TRPO~\cite{schulman2015trust}, then use it to generate  sub-optimal demonstrations (indicated as Demonstration in Table~\ref{tab:result}) including only one trajectory with a Gaussian noise added to each action. In order to compare across the methods, all policies and value function are parameterized by fully connected neural network with two hidden layers consisting 64 nodes each and activated by $\tanh$ function. Also all methods are evaluated within same amount interactions with environment.

The methods we compare LEARN-SAM to include
\begin{itemize}[leftmargin=9pt]
    \item pre-training \cite{silver2016mastering}: We train an expert policy given the demonstration and then use it as the initialized policy for the optimization of the policy;
    \item shaping \cite{brys2015reinforcement}: Reward shaping;
    \item POfD \cite{kang2018policy}: We use a fully connected NN with two hidden layers with 64 and 32 nodes, respectively, for the classifier;
    \item TRPO (Eqn \eqref{eqn:TRPO});
    \item MMD-IL \cite{kim2013maximum}. We enable TRPO when the agent steps into states with large MMD measure instead of applying instructions from an online oracle policy. This method verifies the reward sparsification.
\end{itemize}
%We compare our method against pre-train method \cite{silver2016masterinvg}, Penalty with reward shaping \cite{brys2015reinforcement}, and POfD \cite{kang2018policy}. In particular, we use Gaussian kernels in potential function for reward shaping, and for POfD, we enabled a fully connected neural network with two hidden layers with 64 and 32 nodes respectively for classifier. Additionally, we verified the reward sparsification by running TRPO and the   sub-optimalness of demonstrations by executing Penalty method with MMD measure \cite{kim2013maximum}. 

For the LEARN-SAM procedure, we used a weighted $l_2$ distance for $d$ of the $\lambda$-function, where the weight  each dimension of the state space is the corresponding inverse SD of the demonstrated states,  and a linear function $\varphi(d) = h d$, where $h$ is a hyperparameter that was tuned for fast learning. We used the softmax for $g$ and set $c=4$ in Eqn \eqref{eqn:psi-softmax} and tuned $b$ in  Eqn \eqref{eqn:psi-softmax} in each experiment to achieve faster learning. We set $\delta=0.01$ in the TRPO optimization in Eqn \eqref{eqn:TRPO} in all 6 experiments. When applying the SAM mechanism in each epoch of the learning process, we set $K=2$, identifying and dividing the mini-policy in each state of the expert policy into either the ``helpful'' or ``unhelpful'' category (that is, $m=1$ - only one expert policy, and $K=2$). To boost learning, the mini-policies that fall in the unhelpful category are not used though the corresponding $\xi_{jk}$ values may not be 0,  %restrict the weight $w'_2$ cannot exceed an upper bound.
leading to violation of the theoretical invariant property in Prop~\ref{prop:splitmerge}. Empirically, we found this has minimal impact on the convergence of the ensemble policy for the following reason.  $\xi_{jk}$ for the mini-policies in the unhelpful category tends to diminish to 0 with epochs. In other words,  even when we optimize the ensemble policy using the exact $\xi_{jk}$ values calculated in each application of SAM, those $\xi_{jk}$ values will eventually approach 0, at a cost of slower learning, upon the convergence of the LEARN-SAM procedure.
%$\|s\| = \|s \oslash \hat{\sigma}\|_2$ where $\hat{\sigma}$ is the standard deviation across the states of the demonstration per the Silverman rule of thumb. in the formulation of the $\lambda$-function;  \cite{silverman2018density} and we used $10((p + 2)n/4)^{1/p})$ ($p=|\mathcal{S}|$ and we used $10((p + 2)n/4)^{1/p})$ ($p=|\mathcal{S}|$  and $n$ is the number of state-action pairs in the demonstration)
%(b= -20, -1.5, -10, -10, -30, -30)/ 

\begin{figure}[!htb]\vspace{-3pt}
\centerline{\includegraphics[width=0.9\textwidth]{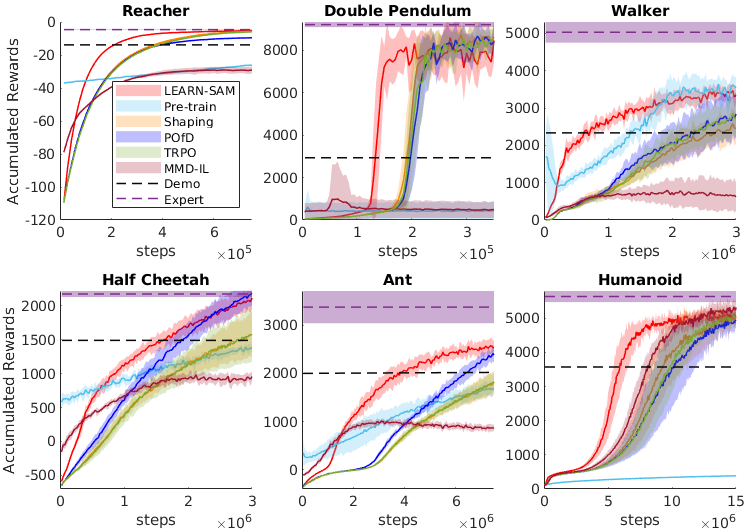}}\vspace{-6pt}
\caption{Accumulated rewards in 6 experiments. Each x-axis represents the number steps that the agent interacts with the environment.}
\label{fig:learningcurve}\vspace{-12pt}
\end{figure}

The  results on the accumulated exact rewards upon convergence  and the iteration number to reach the accumulated exact reward achieved by the demonstration  (without discounting) over 10 repeats are presented in Table~\ref{tab:result}. The trajectories of the mean accumulated rewards over the number of steps are  displayed in Figure~\ref{fig:learningcurve}. The results indicate that LEARN-SAM delivers superior or similar performance  compared to its competitors and reaches the accumulated exact reward level achieved by the demonstration the fastest with significantly fewer steps. 
\begin{table*}[!htb]
\caption{Mean  $\pm$  standard deviation of the accumulated exact reward and the iteration number to the accumulated exact reward achieved by the demonstration over 10 repeats in 6 experiments (the best performance is in bold, and the second best is in italics).} \label{tab:result}\vspace{-12pt}
\begin{center}
\resizebox{1.0\linewidth}{!}{
\begin{tabular}{@{} c@{\hspace{6pt}}c@{\hspace{6pt}}c@{\hspace{6pt}} c@{\hspace{6pt}} c@{\hspace{6pt}} c@{\hspace{6pt}} c@{}}
\hline
    &  DoublePendulum-v2   & Reacher-v2   & Walker2d-v3    & HalfCheetah-v3   & Ant-v3   & Humanoid-v3 \\
\hline
cardinality $|\mathcal{S}| / |\mathcal{A}|$ & 11 / 1 & 11 / 2 & 17 / 6 & 17 / 6 & 111 / 8 & 376 / 17\\
reward setting & Type 3 & Type 1 & Type 2 & Type 2 & Type 2 & Type 2\\
\hline
expert$^\dagger$ & 9221.06 $\pm$ 121.94 & -4.62 $\pm$ 0.14 & 5028.66 $\pm$ 276.12 & 2172.54 $\pm$ 44.24 & 3378.80 $\pm$ 334.46 & 5634.55 $\pm$ 161.99\\
demonstration$^\ddagger$ & 2936.01 & -13.88 & 2333.29 & 1489.77 & 2002.23 & 3570.80\\
\hline
\multicolumn{7}{c}{Aggregated reward at the last iteration} \\
\hline
TRPO & \textit{8549.00} $\pm$ 592.46 & -5.95 $\pm$ 0.18 & 2723.89 $\pm$ 532.70 & 1577.79 $\pm$ 398.24 & 1827.57 $\pm$ 215.65 & 5030.34 $\pm$ 164.14\\
MMD-IL & 490.99 $\pm$ 391.53 & -29.12 $\pm$ 2.05 & 619.91 $\pm$ 378.41 & 957.00 $\pm$ 81.31 & 872.16 $\pm$ 63.42 & \textbf{5315.94 $\pm$ 200.49}\\
Pre-train & 436.95 $\pm$ 423.23 & -26.18 $\pm$ 0.43 & \textbf{3541.81 $\pm$ 278.94} & 1370.00 $\pm$ 31.16 & 1684.07 $\pm$ 125.11 & 387.21 $\pm$ 5.62\\
Shaping & \textbf{8644.67 $\bs\pm$ 385.02} & \textit{-5.74 $\pm$ 0.21} & 2450.38 $\pm$ 414.60 & 1579.14 $\pm$ 329.50 & 1825.08 $\pm$ 227.39 & 5019.61 $\pm$ 161.16\\
POfD & 8416.60 $\pm$ 783.33 & -9.53 $\pm$ 0.39 & 2837.69 $\pm$ 590.43 & \textbf{2183.24 $\pm$ 180.77} & \textit{2381.09 $\pm$ 158.45} & 4922.24 $\pm$ 235.15\\
LEARN-SAM & 8471.40 $\pm$ 1042.73 & \textbf{-5.18 $\pm$ 0.30} & \textit{3356.44 $\pm$ 193.00} & \textit{2113.63 $\pm$ 145.46} & \textbf{2536.66 $\pm$ 135.59} & \textit{5044.19 $\pm$ 185.22}\\
\hline
\multicolumn{7}{c}{Number of iteration to reach the aggregated reward achieved by the demonstration $^*$} \\
\hline
TRPO & 20.20 $\pm$ 1.14 & 39.30 $\pm$ 0.63 & 243.89 $\pm$ 95.69 & 316.50 $\pm$ 128.16 & $>$750.00 & 952.5\\
MMD-IL & 45.55 $\pm$ 14.07 & $>$150.00 & $>$375.00 & $>$500.00 & $>$750.00 & 801.50 $\pm$ 43.14\\
Pre-train & $>$50.00 & $>$150.00 & 88.89 $\pm$ 64.23 & 317.50 $\pm$ 172.80 & $>$750.00 & $>$1750.00\\
Shaping & \textit{19.50 $\pm$ 0.85} & \textit{37.20 $\pm$ 0.63} & 246.39 $\pm$ 57.45 & 297.50 $\pm$ 107.62 & 737.50 $\pm$ 29.37 & \textit{916.50 $\pm$ 149.24}\\
POfD & 20.45 $\pm$ 1.12 & 40.35 $\pm$ 1.12 & \textit{229.72 $\pm$ 68.89} & \textit{189.00 $\pm$ 21.74} & \textit{635.50 $\pm$ 36.17} & 980.50 $\pm$ 167.56\\
LEARN-SAM & \textbf{14.00 $\pm$ 0.58} & \textbf{22.05 $\pm$ 0.73} & \textbf{63.06 $\pm$ 11.44} & \textbf{154.50 $\pm$ 21.88} & \textbf{362.50 $\pm$ 53.14} & \textbf{592.00 $\pm$ 57.07}\\
\hline
\end{tabular}}\vspace{-2pt}
\resizebox{1.0\linewidth}{!}{
\begin{tabular}{l}
\scriptsize $^\dagger$ the  expert policy used to generate  demonstration.$\mbox{\hspace{3in}}$\vspace{-5pt}\\
\scriptsize  $^\ddagger$ the numbers represent the quality of the demonstrations.\vspace{-5pt}\\
\scriptsize $^*$ all iteration numbers are in units of $10,000$ steps.\\
\hline
\end{tabular}}\vspace{-12pt}
\end{center}
\end{table*}

\begin{figure}[!htb]
\centering\vspace{-6pt}
\includegraphics[width=0.8\columnwidth, trim=0cm 0cm 0cm 0cm, clip] {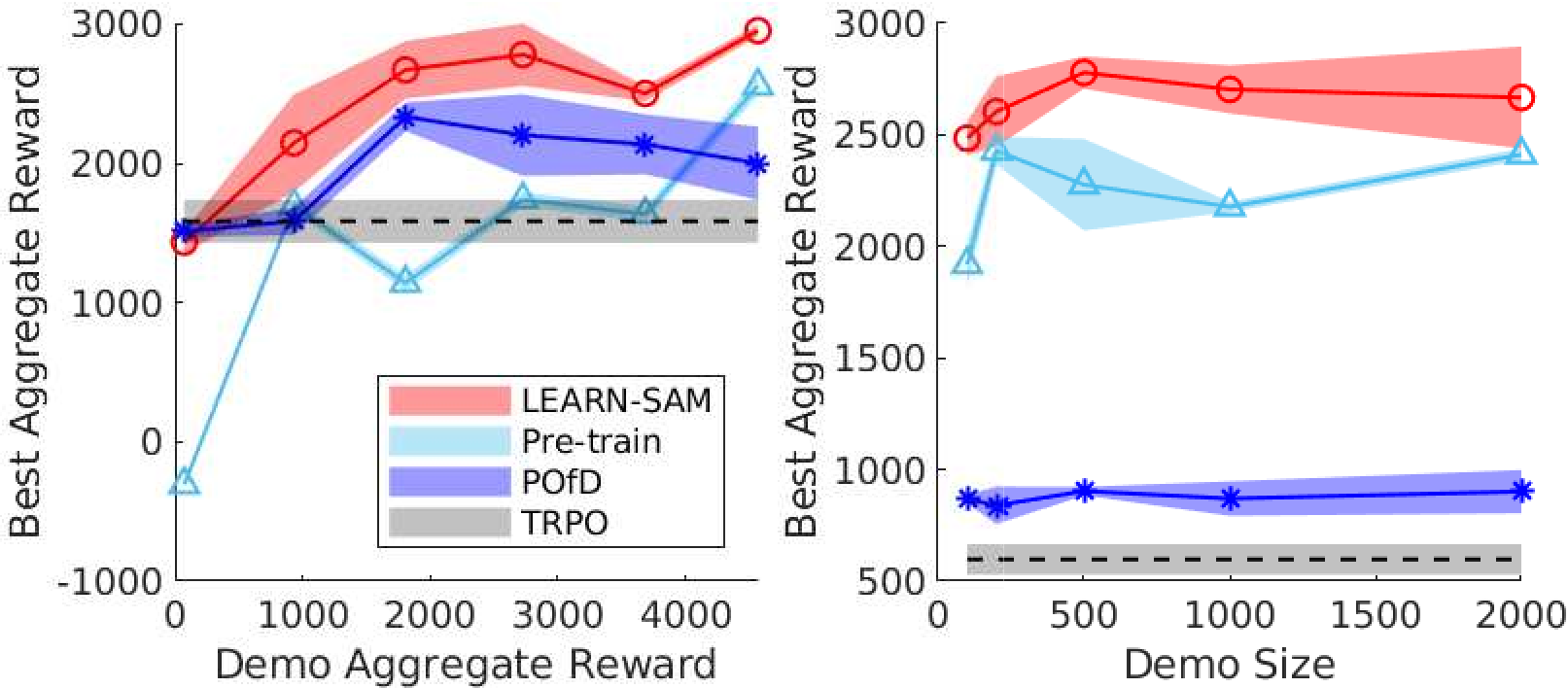}\\
\footnotesize{(a) sensitivity to demo quality \hspace{3cm}(b) sensitivity to demo sparsity}
\vspace{-6pt}
\caption{Sensitivity analysis in the Ant-v3 experiment.}
\label{fig:sensitivity}
\end{figure}

We also run two sensitivity analyses in the Ant-v3 experiment and examine the robustness of LEARN-SAM to the quality and data sparsity of the demonstration, relative to pre-training,  POfD, and TRPO. The reason for choosing the 3 methods  as baseline for the sensitivity analysis is that pre-training and  POfD  perform better than the other benchmark methods in Fig.~\ref{fig:learningcurve} whereas TPRO is used to show the performance without demonstration. The results are presented in Fig~\ref{fig:sensitivity}. 
Specifically, Fig~\ref{fig:sensitivity}(a) shows how performance changes when the demonstration quality varies. The results suggest LEARN-SAM delivers good  performance even when the demonstration quality is relatively low (e.g, $<2,000$ aggregate reward) and reaches a rather stable state when quality is $>2,000$ (with $6.25\times 10^6$ interactions with the environment). Comparatively, pre-training is greatly influenced by the demonstration quality -- with a big jump from 0 to 1,000 around demonstration quality $=4,000$; POfD is somewhere in between LEARN-SAM and  pre-training. 
Fig~\ref{fig:sensitivity}(b) shows the sensitivity to data sparsity of the demonstration. LEARN-SAM outperforms pre-training and POfD within $3.75\times 10^6$ steps across the whole range (100 to 2,000) of the examined state-action pairs of the demonstration. Although pre-train and POfD are also relatively insensitive to the demonstration size, they collect significantly less reward. Since TRPO does not utilize the demonstration information, its performance remains constant regardless of the quality or sparsity of the demonstration in Fig~\ref{fig:sensitivity}(a) and Fig~\ref{fig:sensitivity}(b). 
%steps: One interaction between agent with environment means one step.  Theoretically, yes, it can be Inf;  Practically, no, we will set maximum number of steps in each run

\begin{figure}[!htb]
\centering
\includegraphics[width=0.75\columnwidth] {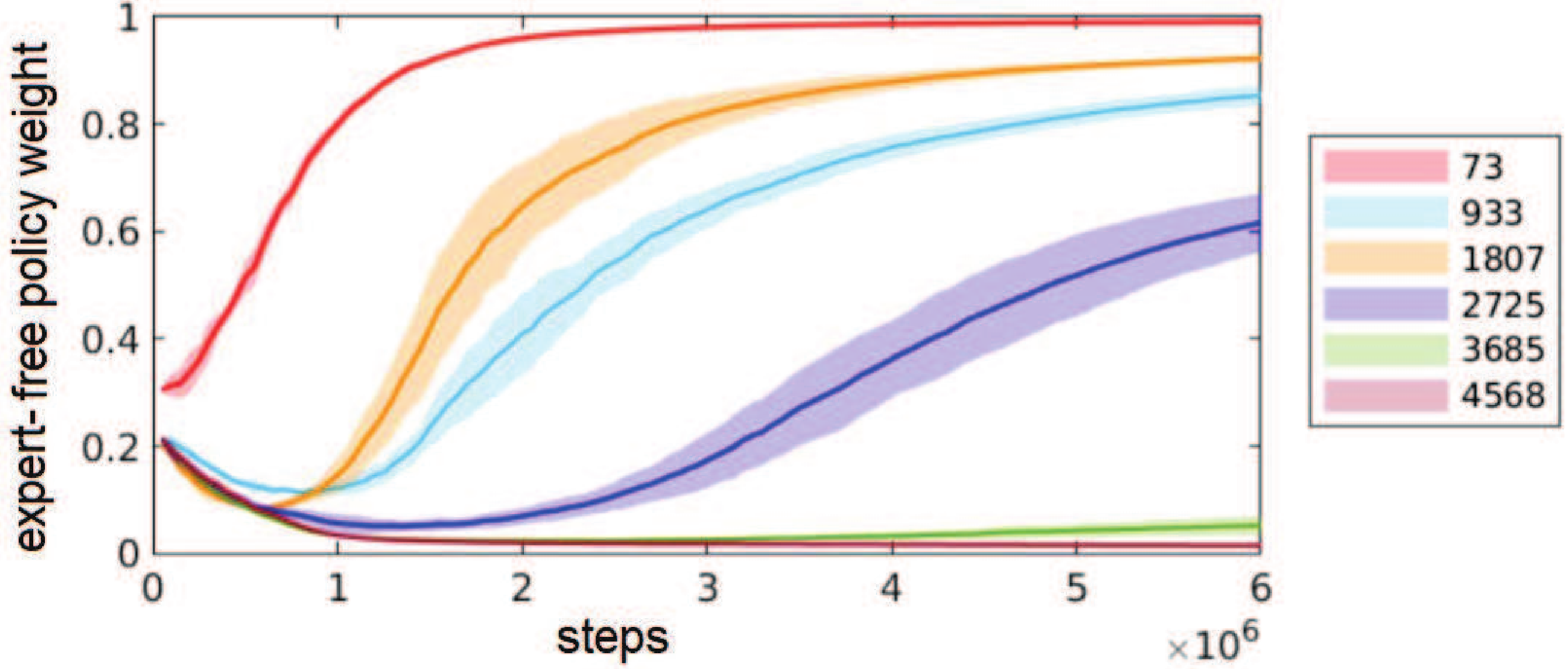}\vspace{-9pt}
\caption{The adaption of the expert-free policy weight $1\!-\!\sum_{j=1}^m\! \lambda_j(s) w_j$ during training and at different demon quality (the numbers in the legends refer to aggregated reward of the demo)in the Ant-v3 experiment.}\vspace{-12pt}
\label{fig:weight}
\end{figure}

Fig~\ref{fig:weight} shows how the local weight of expert-free policy $1\!-\!\sum_{j=1}^m\! \lambda_j(s) w_j$ changes during training  at varying demonstration quality in the Ant-v3 experiment, aiming to show how LEARN-SAM works by examining how the weight of the expert-free policy ``automatcally'' adapts to the demonstration quality during training. Specifically, with low-quality demonstrations, the expert-free policy does not rely on the demonstrate and rather explore the state space on its own; with high-quality demonstrations, LEARN-SAM knows it and makes the best use of the demonstration in learning. %This indicates the amount of information extracted by LEARN-SAM increases with the quality of demonstrations.

\section{Discussion}
We proposed LEARN-SAM to leverage sub-optimal demonstrations in RL in a more effective manner than existing methods for RLfD. LEARN-SAM has two key elements:  the $\lambda$-function that ``localizes'' the weights of  expert policies in each state and the SAM mechanism that reparameterizes  expert policies to make selective usage of demonstration by state.  Between the two, the $\lambda$-function plays a more critical role than the SAM mechanism in term of LEARN-SAM's performance. The SAM mechanism is more of add-on in the sense that it would not be meaningful to apply SAM if there was no $\lambda$-function in the first place. %(we run LEARN without the SAM mechanism in the Walker experiment; its performance  decreases but only slightly upon convergence compared to LEARN-SAM). 
When demonstrations are sub-optimal, the experiments suggest that LEARN-SAM achieves an aggregated reward target much quicker than existing RLfD method.  When demonstrations are of good quality with sufficient data, LEARN-SAM is not necessarily faster than the RLfD methods that assume optimality of the demonstrations though it can still be used %by manually setting initial weights of the expert policies to values close to 1. The experiment suggests that LEARN-SAM spends several epochs evaluating the quality of demonstrations and adjusting the weights of expert policies accordingly.

The $\lambda$ function (but not the SAM mechanism in general) may also be realized in the option-based hierarchical RL framework. Specifically, the option terminates an acting expert sub-policy and then activates a new one with a probability proportional to the weight in Eqn~\eqref{eqn:ensemblelambda} in every step. However, without the help of a formulation similar to the $\lambda$-function in LEARN-SAM that instructs the agent to execute which expert policy (e.g., large/small probabilities if the current state is near/far from the demonstrated states in the LEARN-SAM procedure), the agent would undergo a significant amount of interactions with the environment to achieve what the $\lambda$-function can do more easily and quickly.

For future work, we aim to improve and expand the LEARN-SAM procedure in several directions. For example, the hyperparameters in the $\lambda$ function are specified pre-training in the current framework. One could instead use  monotonically increasing functions, the parameters of which  are estimated simultaneously with the weights and policy parameters during training,  such as polynomials with positive coefficients or monotonic NNs \cite{sill1997monotonic,wehenkel2019unconstrained}. Along the same line, the scoring function and its hyper-parameters $c$ and $b_k$ can also be made adaptive during the training, leading to a more flexible SAM mechanism. We also plan to look into the SAM mechanism more thoroughly such as more ways to define the scoring function and examining empirically how different choices of $K$ and demonstration of various quality affect the performance of LEARN-SAM.

\bibliographystyle{IEEEtranN}
\bibliography{ref}

\end{document}